\documentclass[letterpaper]{article} % DO NOT CHANGE THIS
\usepackage{aaai23}  % DO NOT CHANGE THIS
\usepackage{times}  % DO NOT CHANGE THIS
\usepackage{helvet}  % DO NOT CHANGE THIS
\usepackage{courier}  % DO NOT CHANGE THIS
\usepackage[hyphens]{url}  % DO NOT CHANGE THIS
\usepackage{graphicx} % DO NOT CHANGE THIS
\usepackage{natbib}  % DO NOT CHANGE THIS AND DO NOT ADD ANY OPTIONS TO IT
\usepackage{caption} % DO NOT CHANGE THIS AND DO NOT ADD ANY OPTIONS TO IT
\usepackage{algorithm}
\usepackage{algorithmic}

\usepackage{newfloat}
\usepackage{listings}
\DeclareCaptionStyle{ruled}{labelfont=normalfont,labelsep=colon,strut=off} % DO NOT CHANGE THIS
\floatstyle{ruled}
\newfloat{listing}{tb}{lst}{}
\floatname{listing}{Listing}
\title{Understanding Political Polarisation using Language Models: \\ A dataset and method}
\author {
    Samiran Gode,\textsuperscript{\rm 1}
    Supreeth Bare, \textsuperscript{\rm 1}
    Bhiksha Raj, \textsuperscript{\rm 1}
    Hyungon Yoo, \textsuperscript{\rm 1}
}
\affiliations{
    Carnegie Mellon University\\
}

\usepackage{bibentry}
\begin{document}

\maketitle

\begin{abstract}
Our paper aims to analyze political polarization in US political system using Language Models, and thereby help candidates make an informed decision. The availability of this information will help voters understand their candidates' views on the economy, healthcare, education and other social issues.  Our main contributions are a dataset extracted from Wikipedia that spans the past 120 years and a Language model-based method that helps analyze how polarized a candidate is. Our data is divided into 2 parts, background information and political information about a candidate, since our hypothesis is that the political views of a candidate should be based on reason and be independent of factors such as birthplace, alma mater, etc. We further split this data into 4 phases chronologically, to help understand if and how the polarization amongst candidates changes. This data has been cleaned to remove biases. To understand the polarization we begin by showing results from some classical language models in Word2Vec and Doc2Vec. And then use more powerful techniques like the Longformer, a transformer-based encoder, to assimilate more information and find the nearest neighbors of each candidate based on their political view and their background. 

\end{abstract}

\section{Introduction}
Polarization among the two main parties in the US, Republican and Democratic, has been studied for a long time (\cite{poole1984polarization},\cite{khudabukhsh2021we}). A lot of the discussion online has become polarized\cite{jiang2020reasoning}, and this discussion gets the most traction online\cite{jiang2020reasoning}. This polarization can affect the decision-making ability of a candidate if selected\cite{chen2022price}. In such scenarios, it is important for users to be able to separate the rhetoric and understand how polar a candidate is. With this work, we set out to ask exactly these questions, "Can we measure how polarizing a candidate is?", "Can we measure how much this polarity has changed over time?", We try to answer these questions using Natural Language based techniques and in the process, create a dataset that will be useful for the research community in trying to understand political polarization in the US. Though we have worked on the US political system, the methods we suggest for measuring polarization would be useful for other countries with similar democratic elections in determining how polazised a candidate is.
We first try classical methods such as Word2Vec\cite{mikolov2013efficient} and Doc2Vec\cite{le2014distributed} to understand if we can find polarization using the data we have and gain more insights. We find that words that are politically sensitive\cite{pew2019politically} are related to other words which are politically sensitive\cite{pew2019politically}. We thus move on to more recent and sophisticated models built using Transformers\cite{vaswani2017attention} to gain more insight into the data. We then use these, in particular, longformers\cite{beltagy2020longformer} to project candidate-specific data into a particular embedding space and then use this data to find the nearest neighbors of each candidate and provide one metric to find how polarized a candidate is. 

% The aim of this project is to gain insights using textual data to understand how different categories of information can help classify the political leanings of an individual. In addition to classification, this project will focus on understanding the rationale of the output and also highlight the features having the highest impact and experiment further with them. We first begin by collecting data from the Wikipedia articles of the politicians, predominantly senators, congressmen etc. Since most of the politicians in the United States belong to one of the two parties, the data collected is already annotated. We plan on using simpler word embeddings such as word2vec, glove and FastText to generate a baseline for classifying politicians based on the data collected from their Wikipedia articles. The metric used will be F-1 score. We hope to understand how choosing certain categorical information affects the classification. For example, will the early life of a politician have any affect on their political polarization and the party they are affiliated with. After achieving the baseline classification, we plan to implement sate of the art attention based models like longformer, DeepMind's percierver, RoBERTa, USE, etc. 

\section{Related Work}
\cite{khudabukhsh2021we} talks about political polarization online and uses machine translation to interpret political polarization on the internet. \cite{bhatt2018illuminating} discusses the impacts of hyper-partisan websites on influencing public opinion as illustrated by their ability to affect certain events in the 2016 US general elections. The authors then go on to show how certain political biases are assumed for the purpose of their study, namely overt support for either a Democrat or Republican is taken to be an indicator of the site being either Liberal or Conservative. This paper is fundamental to our research as it looks into the political division and lays the foundation for any following work in the domain of using specific features to classify an entity as being Liberal or Conservative. The features they considered were transcripts of the content being published or shared on these sites. In our case, the features will simply be the Wikipedia page content of the people. \cite{khudabukhsh2022fringe} shows the polarization in TV media and fringe new networks and uses Language model-based techniques to understand them further. However, this polarization visible in the electorate stems from the candidates.
\cite{desilver2022polarization} claims that the candidates become polarized and moved away from the center over the years. With this paper, we release a dataset and a few metrics that will help us understand if political polarization exists in political candidates and how we might be able to measure this political polarization. The aim of this study is to aid voters to make informed decisions before elections. And we use language-based techniques on a dataset that is classified into 4 eras and divided into 3 parts, mainly background, political and other.
% In \cite{zaidan2007using}, the paper discusses the impact of annotator rationales' on certain aspects of training data, namely the use of richer training data. This approach shows that effective training is possible even when the amount of data is relatively less.
% The authors have shown that this approach can help in sentiment analysis. They used discriminative SVM based methods to reach their conclusion.
% In our case we have a very small data set. Hence the annotations are used to elucidate the role of a human as a teacher. In our model, we have annotated the data to broadly classify it into 3 categories. 'Background','Political Career' and 'Others'. This is a type of teaching mechanism that is explained by the paper. As indicated in the paper, we will not be exclusively reliant on the annotations, but just use them as extra information.

\cite{belcastro2020learning} Demonstrates that Political Polarization can be mapped with the help of Neural Networks. This is almost a baseline idea as we are using attention networks and Longformer models for the same. The key difference lies in the data extraction and methodology.

\cite{khadilkar2022gender} goes in depth towards finding gender and racial bias in a large sample of Bollywood (and Hollywood) movies. The author has amalgamated several known NLP models while he tries to create a reasonably robust model of his own. The portions in which this particular study differs from those before is that the sample size is fairly large. It then diverges further with its rather innovative use of diachronic-word embedding association tests (WEAT). Other techniques that are implemented include count-based statistics dependent on a highly popular lexicon cloze test using BERT as a base model (an idea we could consider after data attention) and bias recognition using WEAT. The final model is a combination of the above three. This paper is highly relevant to our project as it uses a similar idea of our own. It uses aforementioned models to predict bias, i.e. sentiment prediction. In our project, we use data to predict political sentiment and attempt to classify certain features as being precursors to classification.

\cite{rajani2019explain} tried to improve speech-based models on their ability to verbalize the reasoning that they learned during training. It uses the CAGE framework (Common-Sense Auto-Generated Explanations) on the common sense explanation dataset to increase the effectiveness by 10 percent. It introduces improvements over the use of BiDAF++ (augmented with self-attention layer) in these newer models. It further uses NLE as rationale generalization within the second phase primarily as means for sentiment analysis. In this paper, Mturk (from Amazon) is used to generate explanations for the dataset.  CAGE primarily uses a question-answer format with 3 options, a label and the best explanation for that label. Furthermore, other evaluation parameters affecting performance are tested and may be used in our project either as verification models or otherwise. CAGE is certainly an interesting choice for verification given the higher accuracy it attains. A factor to be considered however is that the types of datasets and models are very different. Thus certain modifications will be made to the above framework.

\cite{devlin2018bert} is the introduction paper for BERT, a model that will be used extensively. It also shows the results of fine-tuning BERT. These indirectly or directly will be used either as pre-trained constraints or as tuning methods.
petroni2019language

\cite{petroni2019language} Demonstrates the ability of pre-trained high-capacity models like BERT and ELMo to be used as knowledge repositories. This is mainly based on 3 observations, (1) The relational knowledge of these models is competitive to that of an NLP with access to certain oracle knowledge. (2) The effectiveness of BERT in an open domain question answer test and (3) The fact that certain facts are easily learnable. The Authors also demonstrate the usage of other models (unidirectional and bi-directional) in the study, namely 'fariseq-fconv' and 'Transformer-XL'. They conclude by showing that BERT-Large is able to outperform other models and compete even with supervised models for the same Task.

\cite{palakodety2020mining} demonstrates the ability of BERT and similar LM's to track community perception, aggregate opinions and compare the popularity of political parties and candidates. This is demonstrative of our work as we intend to use BERT for the purpose of sentiment analysis. The authors conclude by stating that the LM can be used as a pipeline for extracting Data in the future.

In \cite{hamilton2016diachronic} the authors try to counter the problem of word meaning changing semantically with context. They propose a robust method by using embeddings. These are then evaluated with the 'Law of Conformity' and 'The Law of Innovation'. These display the role of frequency and polysemy in the building structural blocks of language. These blocks will be crucial for 2 reasons, (1) The meaning changes may adversely affect sentiment analysis and thus affect results. Thus frequency and polysemy must be duly curtailed.
(2) The embedding research is fundamental as we are using embedding-based models. Specifically Word2vec.

\section{Dataset Description}

\subsection{Source}
Our data is sourced from the individual pages of politicians(Senators and Congress members) from the 58\textsuperscript{th} to the 117\textsuperscript{th} congress. We divide these into 4 phases, chronologically, with each phase consisting of about 14 congresses. For each congress member, we scrape the section-wise data.

% The Wikipedia pages of all the politicians(Senators, Congressmen, etc) already contain annotated text that can be used for this project with minimal pre-processing. Hence, our primary dataset would consist of personal Wikipedia pages of all the politicians we aim to study. We believe that some sections like early life and political career will have the most weight in deciding the polarization. We hope to understand the impact of other sections of the data as well.

\subsection{Data Collection and Processing}
We scrape Wikipedia based on the list of politicians from the Wikipedia page for each congress. For each congress member in the list, we store the label, their party and the metadata. For each instance, this includes their personal details and all the information from their page as a dictionary, with the heading being the keys and the content being the value. This information helps with the downstream task of cleaning. We annotate this data based on the experiment, in our case we have manually annotated the data to classify these keys into three separate categories. 1) Background data, 2) Political data, 3) Other; in our release, we will be releasing both the annotated and raw versions to facilitate custom use.
% Need to add statistics for the data across the phases. Number of politicians in each phase, text length across categories, and images would be best. 
Wikipedia page sections don't have a fixed format, each politician has different key sections. For instance, Early Life and Background can be split into many sections such as Education, Career, Family, Personal History, etc. So all these sections are grouped into a single category Background. Similarly, anything related to their political affiliation, elections, campaigns and positions held during their tenure are categorized into a single annotation Political Career. All other categories such as Awards, Controversies, Business related activity, Post political career are clubbed under the Others category. This way, only relevant data is selected under each category by manually changing the annotation based on the content inside each category. To conclude, for just Phase 4, a total of 1656 categories were merged into 3 categories for 1631 instances in the first pass spread over roughly 26 years(1995-2021).
This data still contains information names, organizations, locations, numbers, etc. which need to be cleaned. We first run a NER model on the data to remove the names and organization. However, we remove location names only from the political section. The reasoning behind this is, to make sure information from the political section is not influenced by location information. However, for background, we want to understand where a person was born and raised affects their political views and for this only this was kept but others were deleted. This information, after the NER, is passed to remove numbers and other irrelevant regular expressions. This makes sure the data being passed for other downstream tasks is clean and gives unbiased answers.
\begin{figure}[h!]
  \centering
  \includegraphics[width=0.5\textwidth]{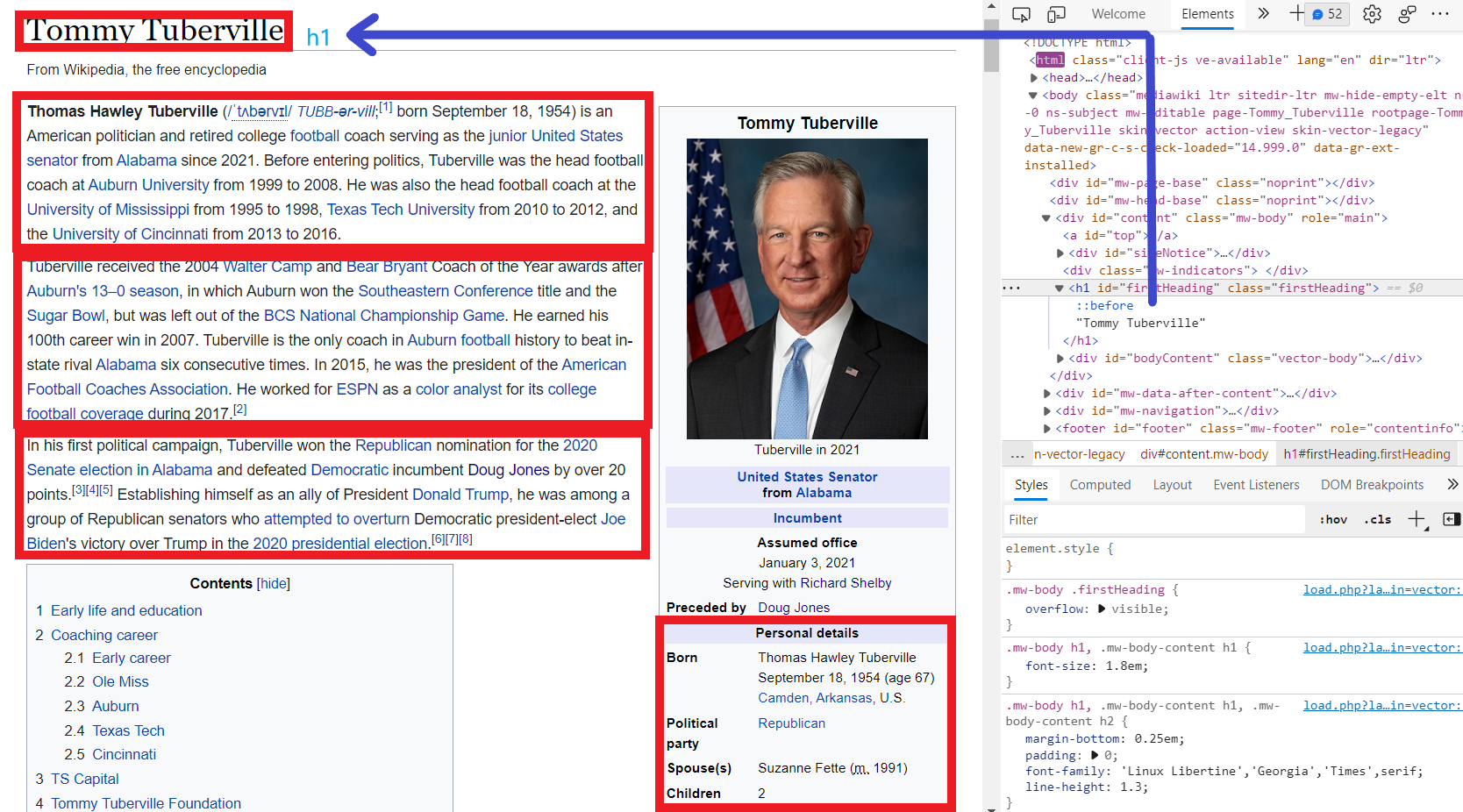}
  \caption{Webscraping based on each Tag}
\end{figure}
\begin{figure}[h!]
  \centering
  \includegraphics[width=0.5\textwidth]{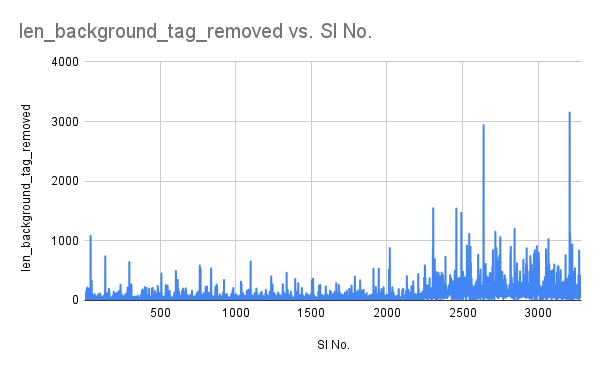}
  \caption{Background}
\end{figure}
\begin{figure}[h!]
  \centering
  \includegraphics[width=0.5\textwidth]{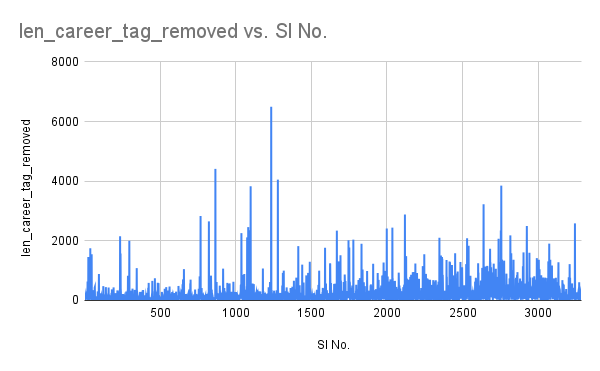}
  \caption{Political}
\end{figure}
\section{Language Model}
Natural Language Processing based applications have been dominated by transformer-based language models where models like BERT\cite{devlin2018bert} and RoBERTa\cite{liu2019roberta} have been state of the art since 2018 but when it comes to our dataset, these models have a drawback, that is, their ability  to process longer sequences since the cost of attention grows on the order of O(N\textsuperscript{2}). Longformer\cite{beltagy2020longformer} and other variants are useful for this task, they accept 4096 input tokens as opposed to 512 for BERT. It reduces model complexity by reformulating the self-attention computation. The performance of Longformer against the current SOTA is represented by the table present below on the raw data.

\section{Experiments}

\subsection{Preliminary Experiments}
Our initial experiments were aimed at gaining insights about patterns or trends that might be present in our data, and also questioning if polarization exists. We do these preliminary experiments using the Doc2Vec\cite{le2014distributed} and Word2Vec\cite{mikolov2013efficient} models. The Doc2Vec model was built from scratch with the raw data, where each Wikipedia page is considered to be a document. We first use the Doc2Vec model with K-means clustering and get a classification accuracy of 59.52\% with political data and 61.846\% with background data. We then used the same Doc2Vec model with binary SVM classifier and achieved an accuracy of 72.872\% with political data and 63.564\% with background data. These results are summarized in the table presented below. The Word2Vec tests were run on pre-trained models as well as models we built from scratch and trained using the data we collected. We used the Word2Vec approach to find approximate nearest neighbors and exact nearest neighbors for certain words on both the Democratic and the Republican sides. This nearest-neighbor approach led to some interesting insights. We expected to see some disparity in the nearest neighbor searches for the Republican data and Democratic data basis the assumption that there is polarization. However using the simple Word2Vec models the 15 nearest neighbors we got were quite similar but as there were certain words for whom the order of the neighbors changed based on the party, for example, for the word 'GUN' , 'VIOLENCE' is the 2$^{nd}$ nearest neighbor(approximate nearest neighbor using spotify's annoy algorithm) for democratic data however the same word is 9$^{th}$ for the republican case, similarly the word 'CHECKS' is the 3$^{rd}$ nearest neighbor for democrats while it is the 8$^{th}$ for republicans. There are more such interesting examples which coupled with the results from the Doc2Vec classification results, prove that political polarization exists and can be learned using Natural Language Processing based techniques. 

%\begin{table*}[t]
%\centering

%\begin{tabular}{l|l|l}
%\textbf{Rank} &
%\textbf{Democrats} &
%\textbf{Republican} \\
%0 &
%gun &
%gun \\
%1 &
%abortion &
%cannabis \\
%2 &
%violence &
%birth \\
%%3 &
%checks &
%lgbt \\
%%4 &
%legalization &
%lgbtq \\
%%5 &
%ban &
%abortion \\
%6 &
%immigration &
%causes \\
%7 &
%background &
%%cell \\
%8 &
%lgbt &
%checks \\
%9 &
%supports &
%violence \\
%10 &
%opposes &
%legalization\\

%\end{tabular}
%}
%\caption{Top 10 nearest neighbors of GUN found using Word2Vec}
%\label{table1}
%\end{table*}

% \begin{table}[h]
%   \caption{Top 10 nearest neighbors of GUN found using Word2Vec}
%   \label{Word2Vec Results}
%   \centering
%   \begin{tabular}{lll}
%     \toprule
%     \cmidrule{1-2}
%     Rank & Democrats  & Republicans \\
%     \midrule
%     0 & gun & gun \\
%     1 & abortion & cannabis \\
%     2 & violence & birth \\
%     3 & checks & lgbt \\
%     4 & legalization &lgbtq \\
%     5 & ban & abortion \\
%     6 & immigration & causes \\
%     7 & background & cell \\
%     8 & lgbt & checks \\
%     9 & supports & violence \\
%     10 & opposes & legalization \\
%     \bottomrule
%   \end{tabular}
% \end{table}

\subsection{Main analysis}
As part of our preliminary analysis, we use RoBERTa, we notice the removal of the words "Democratic", "Republican" etc. causing a drop in classification. This is expected as we lose obvious information and classifying just based on the first 512 tokens is challenging. We hence use Longformer since it can consider 4096 tokens at a time. As expected, this increases the score significantly, as can be seen in the table. There are two versions of Longformer - "longformer-base-4096" and "longformer-large-4096. Longformer base provides a significant improvement over the previous model RoBERTa and simpler models such as Doc2Vec. However Longformer large provided a better score and has been the best performing model when it comes to classifying a given candidate's political party. We also analyze using BigBird which is yet another model which can consider tokens with lengths of 4096, the results of these experiments are in the table. We use this information to understand how different scores are affected by different words and relate this with our broader aim of Political polarization. For that, we calculate the global attention scores of the last layer and then find the words that have the highest attention scores for self-attention with the  $<s>$ token.  This has shown some interesting results, for example, for Ted Stevens, a Republican, some obvious words like, "public", "federal", "legislature", "Wisconsin" show up higher which is expected since the main information from the political text is related to their work, but the word "abortion" showed up in the top 10 percentile words, more such analysis is being done which we believe will give more interesting results, the above analysis for simpler models like BERT isn't as impressive since the information is local and for Longformer this not trivial since Longformer looks as context using sliding windows, however, the Longformer architecture allows for certain tokens to have global attention and choosing the CLS token allows us to look at the attention of all 4096 tokens with this word. Another hypothesis that we have been testing is that the background of a candidate can also help us identify the political leaning of a person which if the world was not polarized would not be the case and only the political information would help us classify, however as can be seen in the table the background matters significantly as well. 

\section{Results}

\begin{figure}[h!]
    \begin{minipage}{0.55\textwidth}
        \includegraphics[width=0.9\textwidth]{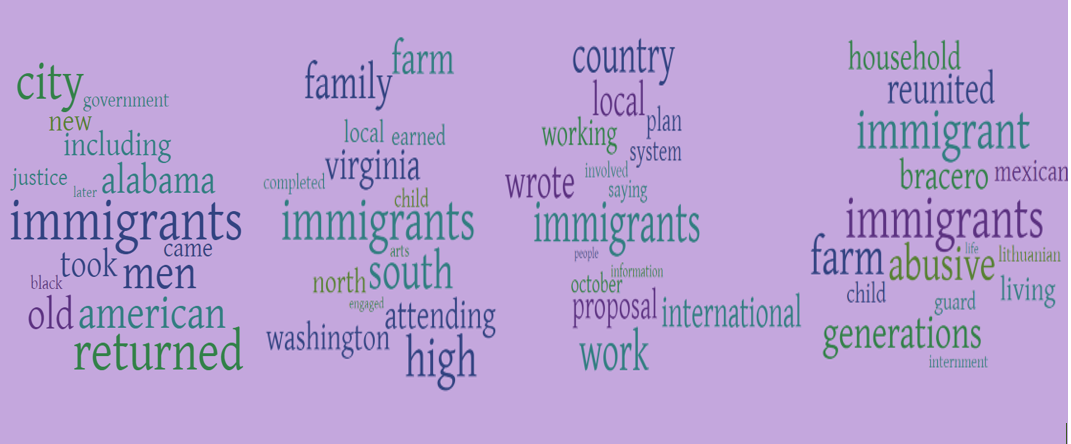}
        \caption{Nearest neighboring words to the word immigrant in the democratic corpus across time from left to right, as we can see, words like americans were closely associated in the early 20th century.}
    \end{minipage}\hfill

\end{figure}

\begin{figure}[h!]
    \begin{minipage}{0.55\textwidth}
        \includegraphics[width=0.9\textwidth]{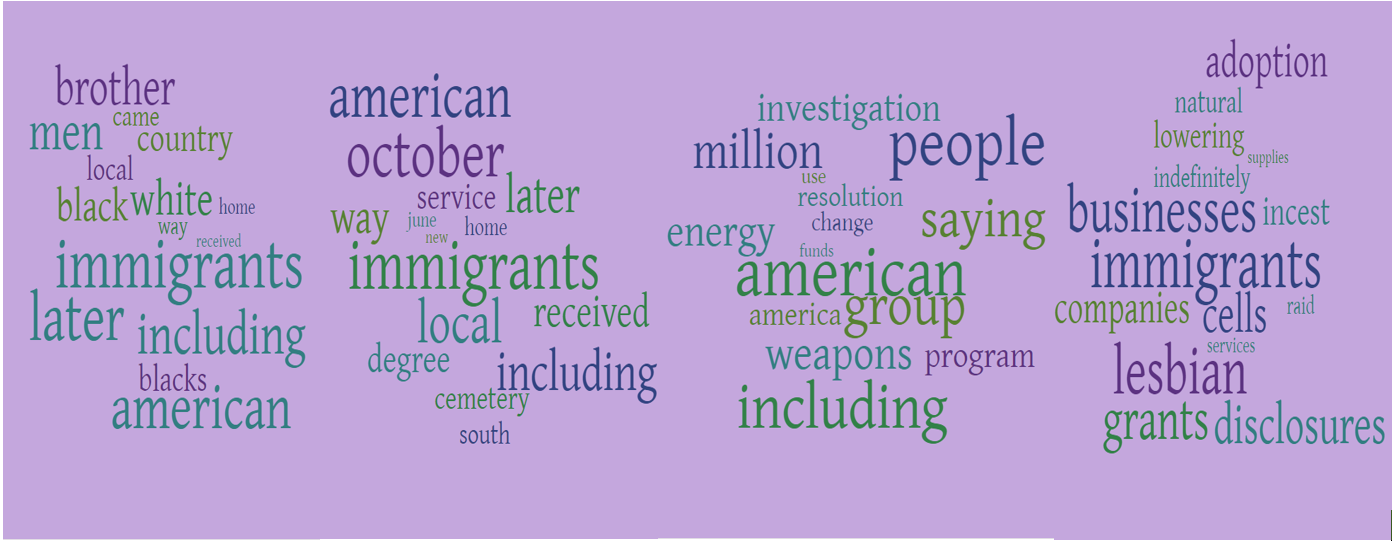}
        \caption{Nearest neighboring words to the word immigrant in the republican corpus, the words are similar in the 1st era, the early 20th century, but in the most recent era it is related to other politically sensitive words.}
    \end{minipage}\hfill

\end{figure}

We tested different models with the annotated raw dataset to understand the polarization in the text. The three models were tested with both the political career dataset and the background dataset to get insight into the factors that influence political polarization. The obtained results are presented in the following table.

\begin{table}[]
\begin{tabular}{lllll}
Model                         & Data       & Accuracy &  &  \\ \cline{1-3}
Doc2Vec                       & Political  & 59.520   &  &  \\
Doc2Vec                       & Background & 61.846   &  &  \\
Allenai/longformer-large-4096 & Political  & 52.128   &  &  \\
Allenai/longformer-large-4096 & Background & 56.383   &  & 
\end{tabular}
\caption{K-means Classification Results}
  \label{K-means Results}
  \centering
\end{table}

% \begin{table}[h]
%   \caption{K-means Classification Results}
%   \label{K-means Results}
%   \centering
%   \begin{tabular}{lll}
%     \toprule
%     \cmidrule{1-2}
%     Model & Data & Accuracy \\
%     \hline
%     \midrule
%     Doc2Vec & Political & 59.520 \%  \\
%     Doc2Vec & Background & 61.846 \%  \\
% %     BigBirdPegasusForConditionalGeneration
% %  & Political & 52.650 \%  \\
% %     BigBirdPegasusForConditionalGeneration
% %  & Background &  57.446 \%  \\
%  %    Allenai/longformer-base-4096
%  % & Political & 52.925 \%  \\
%  %    Allenai/longformer-base-4096
%  % & Background &  53.723 \%  \\
%     Allenai/longformer-large-4096
%  & Political & 52.128 \%  \\
%     Allenai/longformer-large-4096
%  & Background & 56.383 \%  \\
%     \bottomrule
%   \end{tabular}
% \end{table}

\begin{table}[]
\begin{tabular}{lllll}
Model                         & Data       & Accuracy &  &  \\ \cline{1-3}
Doc2Vec                       & Political  & 72.872   &  &  \\
Doc2Vec                       & Background & 63.564   &  &  \\
Allenai/longformer-large-4096 & Political  & 76.862   &  &  \\
Allenai/longformer-large-4096 & Background & 69.681   &  & 
\end{tabular}
\caption{Binary SVM Classification Results}
  \label{Binary SVM Classification Results}
  \centering
\end{table}

% \begin{table}[h]
%   \caption{Binary SVM Classification Results}
%   \label{Binary SVM Classification Results}
%   \centering
%   \begin{tabular}{lll}
%     \toprule
%     \cmidrule{1-2}
%     Model & Data & Accuracy \\
%     \hline
%     \midrule
%     Doc2Vec & Political & 72.872 \%  \\
%     Doc2Vec & Background & 63.564 \%  \\
% %     BigBirdPegasusForConditionalGeneration
% %  & Political & 67.281 \%  \\
% %     BigBirdPegasusForConditionalGeneration
% %  & Background & 62.234 \%  \\
%     Allenai/longformer-large-4096
%  & Political &  76.862 \%  \\
%     Allenai/longformer-large-4096
%  & Background &  69.681 \%  \\
%     \bottomrule
%   \end{tabular}
% \end{table}

Apart from these accuracy tests, we also leverage the attention mechanism of the Longformer model. We find the words with the highest attention scores to correlate them with our theory of political polarization. We have also designed an interactive website that helps you to understand if polarization exists. The website finds the nearest neighbors of the selected politicians from the Longformer output. Then depending on the ratio of Republicans to Democrats in the nearest neighbors, we estimate the politician's polarization. One such example is shown below-

\begin{figure}[h!]
    \centering
    \begin{minipage}{0.45\textwidth}
        \centering
        \includegraphics[width=0.9\textwidth]{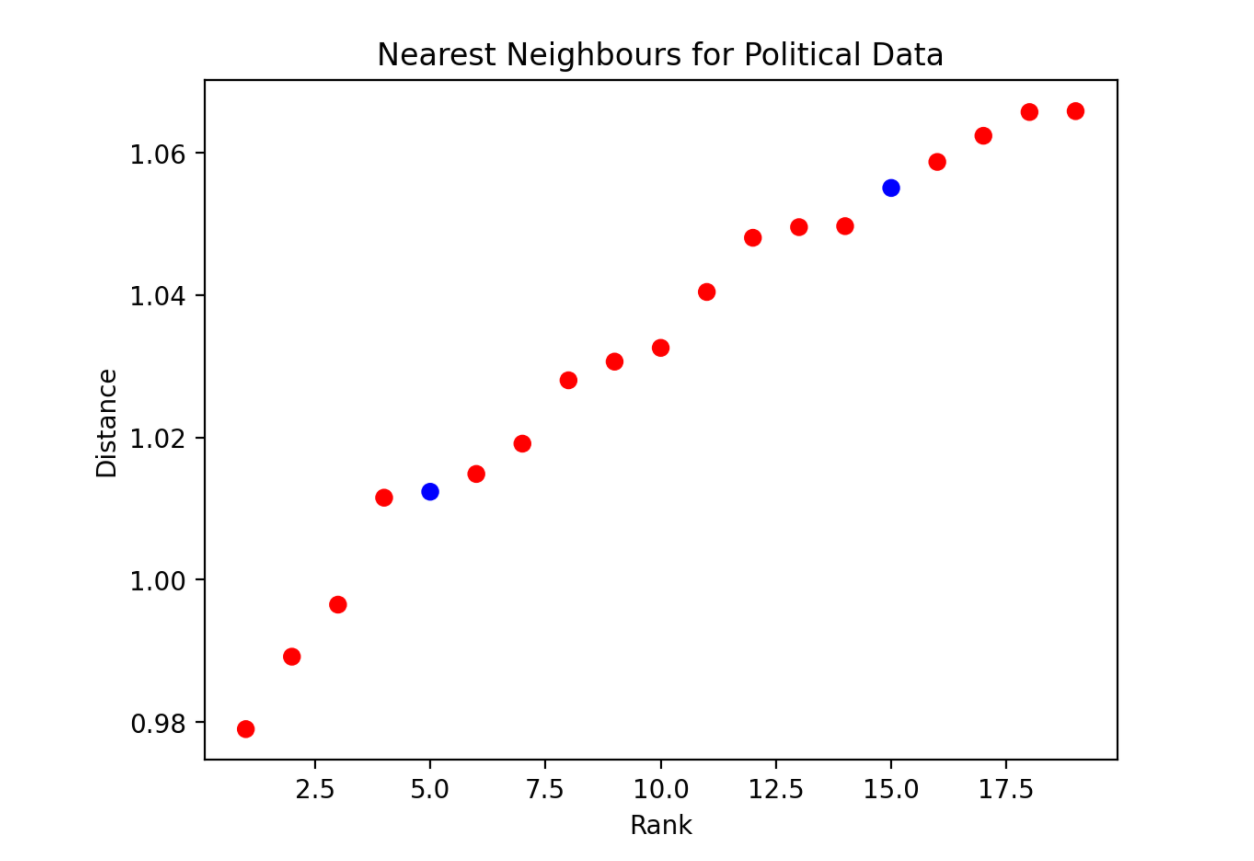} % first figure itself
        \caption{Nearest neighbors for political data belonging to Mitch McConnell (Republican)}
    \end{minipage}\hfill
\end{figure}
\begin{figure}[h!]
    \begin{minipage}{0.45\textwidth}
        \centering
        \includegraphics[width=0.9\textwidth]{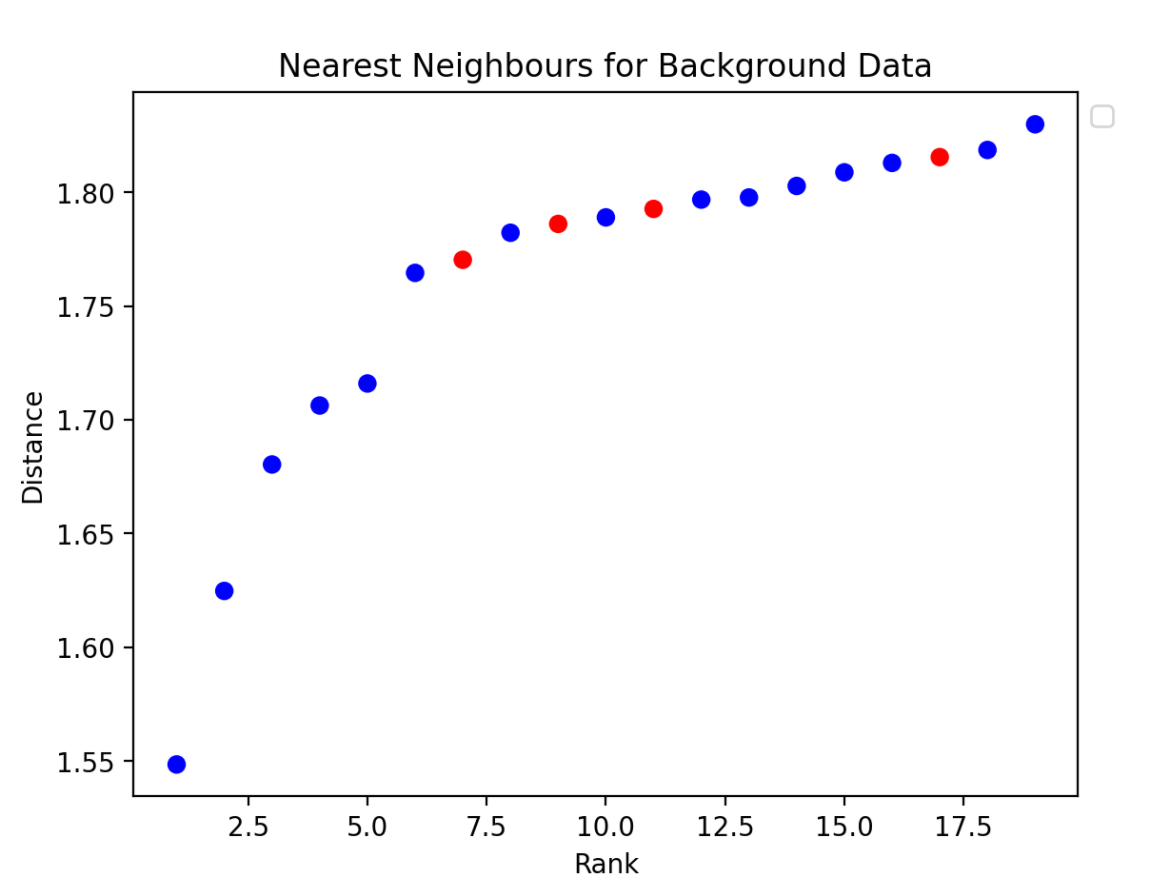} % second figure itself
        \caption{Nearest neighbors for background data belonging to Ayenna Presley (Democrat)}
    \end{minipage}
\end{figure}

In the graph, the x-axis is the rank of the 20 closest neighbors for the politician you choose given the dataset and the y-axis shows their respective closeness scores. The color blue is for Democrats and red for Republicans. The ratio is Blue vs Red points in this graph, so one of our hypotheses is that if a politician isn't polarized this ratio should be 0.5(democrat/total) if we just look at the background data. The above graph in Figure 10 shows the neighbors for Mitch McConnell (Republican) and it is very evident that majority of the neighbors are Republican (red in color) whereas the Democrat count is only 2 out of 20. So one can infer that the polarization ratio is 0.9 for Mitch McConnell. Similarly, in Figure 11, we can see that Ayenna Presley who is a Democrat has 16 neighbors belonging to the same party resulting in a ratio of 0.8 for background data. Scaled-up versions of such websites with more metrics that highlight the political views of a member as radical or moderate will be beneficial to the voters.\\
Also, we show the worthiness of this data and hope that this will be useful to the research community in examining the idea of political polarization in candidates and how it is linked to other attributes of the candidate. Also, to understand the views of a candidate and measure how polarizing their views are. We also hope that the spread of this data across multiple decades will help us understand how political ideas have changed over time.

\section{Future Work}
For future work, we aim to use other metrics for finding the political polarization of individuals and communities again using the Wikipedia dataset. Specifically, we want to use the attention tokens mentioned above to look at the ratio of tokens from the background to the political given text from a candidate that is equally distributed across the background and political. 
\section{Acknowledgments}
We would like to thank Yash Jain and Viraj Ranade for their contributions.

\bibliography{aaai23}

% \begin{thebibliography}{9}
% \bibitem{texbook}
% Donald E. Knuth (1986) \emph{The \TeX{} Book}, Addison-Wesley Professional.

% \bibitem{lamport94}
% Leslie Lamport (1994) \emph{\LaTeX: a document preparation system}, Addison
% Wesley, Massachusetts, 2nd ed.
% \end{thebibliography}

\end{document}